\documentclass[sigconf]{acmart}

\usepackage{booktabs}
\usepackage{colortbl}
\usepackage{placeins}
\usepackage{pdfpages}
\definecolor{groupgray}{gray}{0.93}
\definecolor{deltbg}{RGB}{255,247,218}
\definecolor{delttext}{RGB}{176,101,0}

\AtBeginDocument{%
  }

\setcopyright{none}
\copyrightyear{2026}
\acmYear{2026}
\acmDOI{}
\acmConference[MM '26]{The 34th ACM International Conference on Multimedia}{November 10--14, 2026}{Rio de Janeiro, Brazil}
\acmBooktitle{Proceedings of the 34th ACM International Conference on Multimedia (MM '26), November 10--14, 2026, Rio de Janeiro, Brazil}
\acmISBN{}
\let\acmoriginalfootnotetextcopyrightpermission\footnotetextcopyrightpermission
\renewcommand\footnotetextcopyrightpermission[1]{%
  \acmoriginalfootnotetextcopyrightpermission{%
    \textbf{Preprint notice.} This is the author-prepared arXiv version of a paper accepted at the 34th ACM International Conference on Multimedia (MM '26), November 10--14, 2026, Rio de Janeiro, Brazil. The technical content is preserved, while conference-specific rights metadata and publisher placeholders are intentionally omitted. This version includes the complete supplementary material as an appendix and may differ in presentation from the final publisher-formatted Version of Record. The source archive is provided to support accessibility and reproducibility. Please cite the final conference version once its DOI and bibliographic record become available. Its pagination therefore differs from the conference submission.%
  }%
}

\begin{document}

\title{RefineSVG: Visual Feedback-Driven Reinforcement Learning for Image-to-SVG Generation}

\author{Shaobo Liu}
\authornote{These authors contributed equally to this work.}
\affiliation{%
  \institution{Shenzhen University}
  \city{Shenzhen}
  \country{China}
}
\email{2410105049@mails.szu.edu.cn}

\author{Feiqiao Mao}
\authornotemark[1]
\affiliation{%
  \institution{Shenzhen University}
  \city{Shenzhen}
  \country{China}
}
\email{feiqiao@szu.edu.cn}

\author{Shuaishuai Zhou}
\affiliation{%
  \institution{Shenzhen University}
  \city{Shenzhen}
  \country{China}
}
\email{2400101044@mails.szu.edu.cn}

\author{Yan Zhan}
\affiliation{%
  \institution{Peking University}
  \city{Beijing}
  \country{China}
}
\email{2401210760@stu.pku.edu.cn}

\author{Weiqi Tan}
\affiliation{%
  \institution{Shenzhen University}
  \city{Shenzhen}
  \country{China}
}
\email{2400101088@mails.szu.edu.cn}

\author{Zhiqiong Lu}
\affiliation{%
  \institution{Shenzhen University}
  \city{Shenzhen}
  \country{China}
}
\email{2410105037@mails.szu.edu.cn}

\author{Zhengping Liang}
\correspondingauthor
\affiliation{%
  \institution{Shenzhen University}
  \city{Shenzhen}
  \country{China}
}
\email{liangzp@szu.edu.cn}

\renewcommand{\shortauthors}{Liu et al.}

\begin{abstract}
We propose RefineSVG, a single-step closed-loop visual feedback framework that enables multimodal large language models (MLLMs) to perform high-fidelity image-to-SVG generation through self-correction. Existing MLLM-based approaches rely on single-pass open-loop inference, where the model receives visual input only once and must generate thousands of SVG code tokens without intermediate verification. This paradigm inevitably leads to geometric drift, error accumulation, and visual hallucination on complex images. RefineSVG overcomes this limitation by invoking an external rendering engine after an initial SVG generation pass to compare the rendered output against the target image. The comparison yields a multi-dimensional visual residual map (Diff-Map) that is fed back to the model as a ReAct-style correction signal, driving a targeted correction step. To support this render-observe-correct interaction, we further introduce an SVG-oriented semantic vocabulary that compresses token sequences by over 52\%. A progressive training pipeline spanning supervised fine-tuning, rejection-sampling cold-start data construction, and end-to-end agentic reinforcement learning aligns the model with closed-loop visual correction. Extensive experiments show that RefineSVG consistently outperforms existing baselines in reconstruction fidelity, structural accuracy, and code efficiency. Code is available at \url{https://github.com/liuxiaobo66/RefineSVG}.
\end{abstract}



\begin{teaserfigure}
  \centering
  \includegraphics[width=0.97\textwidth]{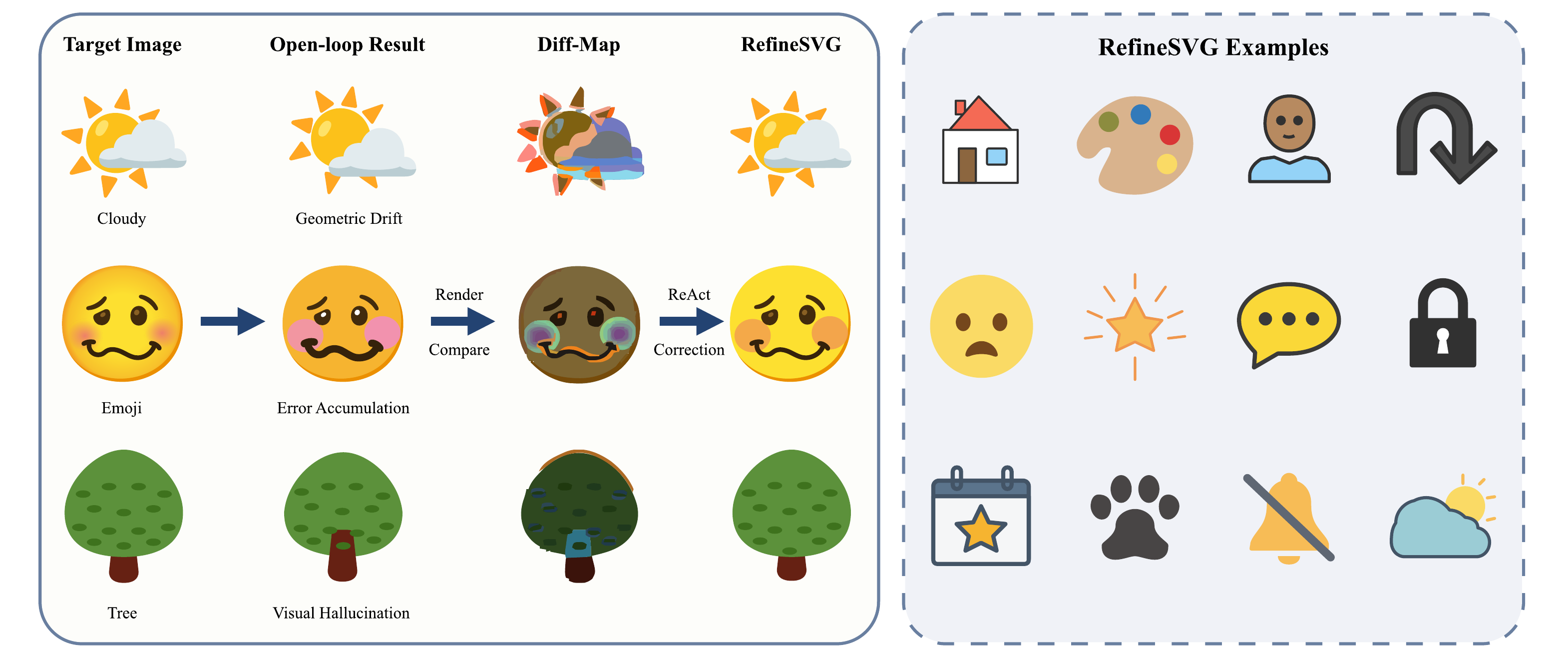}
  \caption{Left: RefineSVG corrects three failure modes of open-loop SVG generation (geometric drift, error accumulation, visual hallucination) via Diff-Map-guided ReAct correction. Right: additional SVG samples generated by RefineSVG-7B.}
  \Description{The left panel shows a target image, an erroneous open-loop SVG rendering, a Diff-Map highlighting visual errors, and the corrected RefineSVG output. The right panel shows twelve SVG icons generated by RefineSVG.}
  \label{fig:teaser}
\end{teaserfigure}

\maketitle

\enlargethispage{-2\baselineskip}

\section{Introduction}

Scalable Vector Graphics (SVG) serves as an essential format in modern digital design, web development, and industrial production owing to its resolution independence, compact representation, and inherent editability. Converting raster images into SVG, commonly referred to as Image-to-SVG generation, has traditionally been approached through optimization-based pipelines. Methods such as DiffVG~\cite{li2020differentiable}, LIVE~\cite{ma2022towards}, and VectorFusion~\cite{jain2023vectorfusion} iteratively refine vector primitives via differentiable rendering or CLIP-guided~\cite{DBLP:conf/icml/RadfordKHRGASAM21} loss functions, producing visually plausible results at the cost of hundreds to thousands of gradient update steps per image. With the rapid advancement of multimodal large language models (MLLMs)~\cite{DBLP:conf/nips/LiuLWL23a, Qwen2.5-VL}, a distinct paradigm has emerged that reframes Image-to-SVG as an inverse rendering and code generation task~\cite{rodriguez2025starvector, yang2025omnisvg}: MLLMs directly read input images and output complete SVG code in a single forward pass, bypassing the iterative optimization loop entirely. This direct generation approach has shown strong performance on simple icons and low-complexity graphics.

However, when confronted with illustrations and complex icons that exhibit intricate geometric structures and rich spatial hierarchies, existing MLLM-based direct generation methods expose severe limitations. The root cause is their single-pass open-loop inference mechanism: the model receives visual input only once and must then produce thousands of code tokens without ever observing the rendered result of its own output. Without mid-stream verification, geometric drift causes coordinates to deviate progressively from intended positions, accumulated errors in color or placement propagate through subsequent elements, and visual hallucination leads the model to produce redundant or contradictory primitives when handling occluded and nested structures (Figure~\ref{fig:teaser}). The model progressively loses visual grounding as the sequence grows, a failure mode that stands in stark contrast to the closed-loop workflow of human designers, who naturally iterate between drawing, observing, and revising.

Motivated by this observation, we propose RefineSVG, a framework for image-to-SVG generation that elevates the MLLM from a passive code generator to a self-correcting visual agent. After the model produces an initial SVG output, the system invokes an external rendering engine to convert the code into a pixel-level preview and compares it with the target image to extract a multi-dimensional visual residual map (Diff-Map) that spatially highlights regions of deficiency and deformation. The target image, the current rendering, and the Diff-Map together form a triplet visual prompt that is fed back to the model, triggering a ReAct correction step in which the model locates and revises the code responsible for the observed discrepancies. This generate-then-correct paradigm enables the model to ground its revisions in concrete visual evidence rather than relying solely on open-loop token prediction.

Realizing this paradigm within MLLM-based code generation poses two technical challenges. First, native SVG code is extremely verbose, and the additional context from visual feedback further inflates the input, rapidly exhausting the context window. Second, existing open-source MLLMs lack the native ability to interpret a visual residual and revise code accordingly, as no such alignment data exists in their pretraining corpora. To address these challenges, we make the following contributions:

\begin{enumerate}
\item We propose \textbf{RefineSVG}, a closed-loop visual feedback framework for Image-to-SVG generation that introduces an external rendering engine\footnote{We use CairoSVG (\url{https://cairosvg.org/}) for SVG-to-raster conversion throughout this work.} and a triplet-based Diff-Map as the correction signal, overcoming the geometric drift and error accumulation inherent in open-loop methods.
\item We construct an \textbf{SVG-oriented semantic vocabulary} with quantized coordinate encoding and prior initialization that substantially compresses SVG token sequences, removing the context bottleneck for closed-loop interaction.
\item We design a \textbf{progressive agentic training pipeline} spanning rejection-sampling-based cold-start data construction and end-to-end reinforcement learning with composite visual rewards, equipping the model with visual correction capability for complex scenes.
\end{enumerate}

\section{Related Work}

\subsection{Optimization-based SVG Generation}

Optimization-based vectorization iteratively refines SVG primitives through differentiable rendering or semantic guidance. DiffVG~\cite{li2020differentiable} enables gradient-based parameter optimization, and LIVE~\cite{ma2022towards} extends it through layer-wise vectorization. CLIPDraw~\cite{DBLP:conf/nips/FransSW22} applies CLIP guidance to stroke optimization. CLIPasso~\cite{vinker2022clipasso} uses related semantic losses for object sketching, while Im2Vec~\cite{DBLP:conf/cvpr/Reddy21} learns vector synthesis without explicit vector supervision. VectorFusion~\cite{jain2023vectorfusion} and SVGDreamer~\cite{xing2024svgdreamer} further use pretrained diffusion models~\cite{DBLP:conf/cvpr/RombachBLEO22} with score distillation~\cite{DBLP:conf/iclr/PooleJBM23}. Although visually effective, these methods require hundreds to thousands of optimization steps and often produce dense, non-semantic path collections with limited topological control.

\subsection{LLM-based Direct SVG Generation}

Early neural approaches represent SVG as structured command sequences. DeepSVG~\cite{DBLP:conf/nips/CarlierDAT20} introduces a hierarchical VAE that encodes path- and command-level structure for generation and interpolation, SVGFormer~\cite{DBLP:conf/cvpr/CaoWE023} learns continuous vector graphics representations via transformers, and IconShop~\cite{DBLP:journals/tog/WuSML23} employs autoregressive transformers~\cite{DBLP:conf/nips/VaswaniSPUJGKP17} conditioned on text prompts. Recent text-to-SVG systems treat SVG synthesis as code generation: SVGen~\cite{wang2025svgen} constructs a million-scale text-SVG dataset and fine-tunes text-only LLMs with curriculum learning, while SVGThinker~\cite{chen2025svgthinker} integrates chain-of-thought reasoning into instruction-aligned generation and editing. These models acquire SVG syntax through task-specific fine-tuning but cannot directly consume raster images. For image-to-SVG generation, StarVector~\cite{rodriguez2025starvector} couples a visual encoder with a code-oriented LLM and contributes the large-scale SVG-Stack dataset and SVG-Bench suite. OmniSVG~\cite{yang2025omnisvg} parameterizes SVG commands as discrete tokens within a vision-language architecture to mitigate coordinate hallucination, while Chat2SVG~\cite{DBLP:conf/cvpr/WuS025} combines LLM-based layout planning with diffusion-based detail synthesis. RLRF~\cite{rodriguez2025rendering} incorporates rendering feedback as a reinforcement learning reward beyond supervised fine-tuning, and VGBench~\cite{zou2024vgbench} supports standardized evaluation. Despite these advances, existing methods remain open-loop at inference time and do not verify their rendered output during generation. RefineSVG instead closes the loop by returning visual residuals to the model as explicit correction signals.

\subsection{Self-Refinement and Agentic Reasoning}

Feedback-driven agents improve LLM outputs beyond single-pass generation. ReAct~\cite{DBLP:conf/iclr/YaoZYDSN023} interleaves reasoning with environment actions, Reflexion~\cite{DBLP:conf/nips/ShinnCGNY23} converts feedback into persistent reflections, and Self-Refine~\cite{DBLP:conf/nips/MadaanTGHGW0DPY23} performs iterative critique and revision. Toolformer~\cite{DBLP:conf/nips/SchickDDRLHZCS23} learns autonomous tool use, while Self-Debug~\cite{DBLP:conf/iclr/ChenLSZ24} and Detikzify~\cite{DBLP:conf/nips/BelouadiPE24} correct programs from execution or compilation feedback. Unlike these text- or code-oriented traces, RefineSVG requires the model to interpret rendered visual discrepancies and revise syntactically constrained SVG markup; we therefore treat the renderer as the external ReAct environment.

\section{Method}
\label{sec:method}

\begin{figure*}[t]
  \centering
  \includegraphics[width=0.98\textwidth]{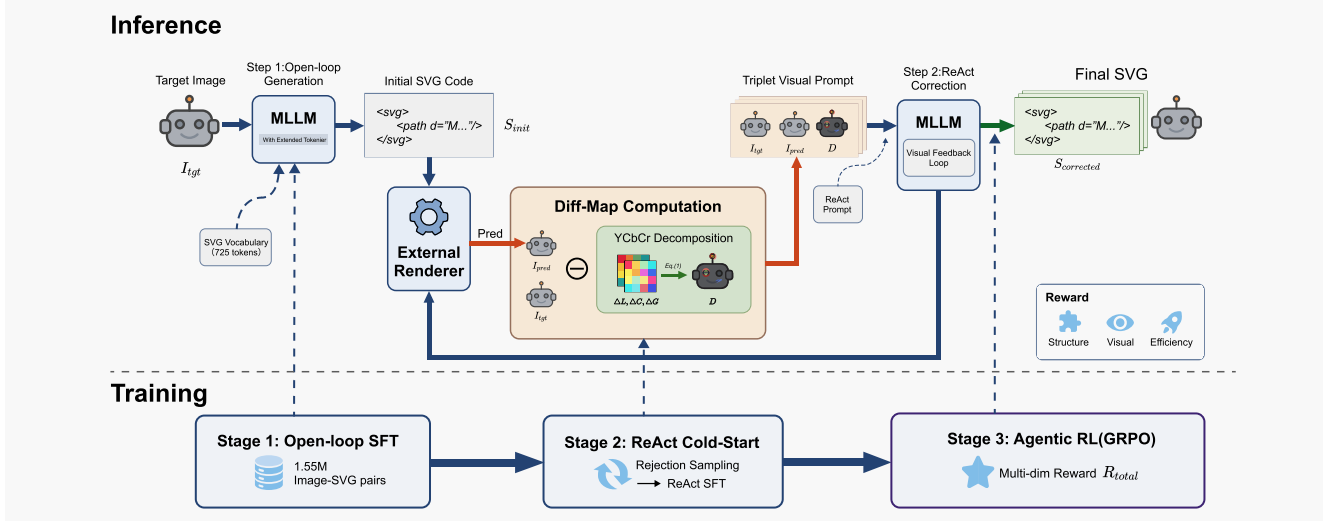}
  \caption{Overview of the RefineSVG framework. \textit{Top}: At inference time, the model first generates an initial SVG from the target image (Step 1: Open-loop Generation). An external renderer produces a pixel-level preview, which is compared with the target via YCbCr decomposition to yield a multi-dimensional Diff-Map. The triplet visual prompt then drives a ReAct-style correction step (Step 2: ReAct Correction). \textit{Bottom}: The three-stage progressive training pipeline: Stage~1 performs open-loop SFT on 1.55M image-SVG pairs; Stage~2 bootstraps closed-loop behavior via rejection-sampling-based ReAct SFT; Stage~3 applies agentic RL (GRPO) with multi-dimensional rewards for structure, visual fidelity, and code efficiency.}
  \Description{A two-part framework diagram. The upper part shows initial SVG generation, external rendering, YCbCr-based Diff-Map construction, and ReAct correction. The lower part shows the three training stages: open-loop supervised fine-tuning, rejection-sampled cold-start training, and agentic reinforcement learning.}
  \label{fig:framework}
\end{figure*}

We propose RefineSVG, a closed-loop framework that introduces external visual feedback into MLLM-based Image-to-SVG generation. As shown in Figure~\ref{fig:framework}, at inference time the model first generates an initial SVG with an extended SVG-oriented vocabulary (Section~\ref{sec:tokenization}), then an external renderer compares the output against the target image to produce a multi-dimensional Diff-Map, which drives a ReAct-style correction step (Section~\ref{sec:react}). The model is trained through a three-stage progressive pipeline: open-loop SFT, rejection-sampling-based cold-start, and end-to-end agentic RL with multi-dimensional rewards (Sections~\ref{sec:training}--\ref{sec:reward}).

\subsection{SVG-Oriented Semantic Tokenization}
\label{sec:tokenization}

The BPE tokenizer~\cite{DBLP:conf/acl/SennrichHB16a} in the backbone MLLM is optimized for natural language and fragments SVG code into sub-word pieces that lack geometric meaning: a tag such as \texttt{<path} may be split across multiple tokens, and floating-point coordinates are arbitrarily truncated, disrupting both structural continuity and spatial accuracy while inflating the sequence length. To bridge this gap, we extend the vocabulary with 725 SVG-specific tokens organized into three categories. Structural primitives encode element tags (\texttt{<svg}, \texttt{<path}, \texttt{<rect}), attribute keywords (\texttt{ d="}, \texttt{ fill="}), path commands (\texttt{M}, \texttt{L}, \texttt{C}, \texttt{Z}), and delimiters (\texttt{/>}, \texttt{</svg>}) as atomic tokens, preserving syntactic integrity at the element level. Quantized coordinates represent spatial positions within a $256 \times 256$ canvas: each coordinate value is decomposed into an integer token (0--256) and a decimal token (\texttt{.00}--\texttt{.99}), so that a coordinate such as \texttt{128.75} maps to exactly two semantically complete tokens. A compact color palette comprises 128 base colors systematically sampled from the grayscale and HSL color spaces, complemented by 128 additional colors obtained by clustering all hex values across the training corpus. To accelerate convergence, we apply semantic prior initialization: each new token embedding is set to the mean of the sub-word embeddings that the base tokenizer produces for a short natural-language description of that token, providing a meaningful starting point instead of random initialization. The complete vocabulary listing and initialization prompts are provided in the supplementary material.

As shown in Figure~\ref{fig:compression}, the extended vocabulary reduces the average token count across over 1.55 million SVG samples from $1977$ to $933$, achieving a compression rate of $52.8\%$. The median (P50) drops from $1248$ to $594$, and the 90th percentile (P90) for complex long-tail samples decreases from $4426$ to $2129$. This compression frees over half of the context budget, alleviating the hallucination tendency in long-sequence generation and enabling closed-loop visual feedback within a bounded context window.

\begin{figure}[t]
  \centering
  \includegraphics[width=0.97\linewidth]{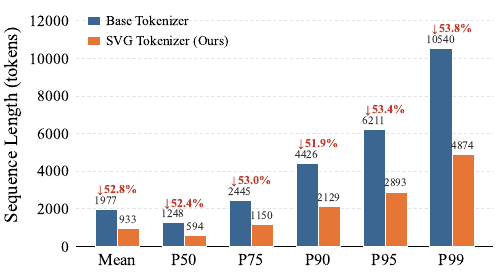}
  \caption{Sequence length statistics before and after applying the SVG-oriented semantic vocabulary. The extended vocabulary achieves consistent compression of over 50\% across all percentiles, from short samples (P50) to complex long-tail cases (P99).}
  \Description{A comparison of SVG sequence-length distributions before and after applying the SVG-oriented vocabulary, showing reductions of more than fifty percent across the reported percentiles.}
  \label{fig:compression}
\end{figure}

\subsection{Visual Feedback-Driven ReAct Mechanism}
\label{sec:react}

After the model produces an initial SVG, the system invokes an external rendering engine to obtain a pixel-level preview $I_{pred}$. To construct a spatially informative correction signal, we compute a multi-dimensional residual between the rendered output and the target image $I_{tgt}$. Both images are converted to the YCbCr color space, and differences are extracted along three perceptually motivated dimensions: a luminance difference $\Delta L$ measuring brightness discrepancy, a chrominance distance $\Delta C$ computed from the Cb and Cr channels, and a structural gradient difference $\Delta G$ that captures edge discrepancies via luminance gradient magnitudes. Each component is independently scaled by a robust percentile-based normalizer $\mathcal{N}(\cdot)$, and the final Diff-Map score is obtained through weighted nonlinear fusion:
\begin{equation}
  S_{diff} = \left(\alpha \cdot \mathcal{N}(\Delta L) + \beta \cdot \mathcal{N}(\Delta C) + \gamma \cdot \mathcal{N}(\Delta G)\right)^{p}
  \label{eq:diff}
\end{equation}
where the weights satisfy $\alpha < \beta < \gamma$ to emphasize chrominance and structural discrepancies over raw brightness following visual perception priors, and the exponent $p < 1$ applies a compressive nonlinearity that smooths transitions while preserving salient error peaks. The continuous score map is converted into a pseudo-color heat map $D$ for visual clarity.

The system concatenates the target image, the current rendering, and the heat map to form a visual triplet $I_{triplet} = [I_{tgt} \oplus I_{pred} \oplus D]$ that is fed to the model alongside a dedicated ReAct prompt. The prompt explains the triplet layout and the color encoding of the heat map, then instructs the model to preserve correct regions of the initial draft while correcting incorrect geometry, colors, structure, and missing elements indicated by the residual. If no meaningful discrepancy is detected, the model returns the draft unchanged, providing a graceful exit condition. The corrected output is wrapped in structured delimiter tags to enable reliable parsing. Detailed parameter values, operator definitions, and the complete ReAct prompt template are provided in the supplementary material.

\subsection{Progressive Agentic Training Pipeline}
\label{sec:training}

The model progresses through three stages: open-loop supervised fine-tuning establishes basic generation capability, rejection sampling constructs cold-start data for interpreting visual feedback, and end-to-end reinforcement learning aligns the full closed-loop behavior.

\noindent\textbf{Stage 1: Open-loop Generation.}\quad
We perform full-parameter supervised fine-tuning on a large-scale image-SVG dataset to equip the model with cross-modal alignment and SVG code generation capability. Given a target image $I_{tgt}$ and its corresponding ground-truth SVG sequence $S^{*} = \{s_1^{*}, s_2^{*}, \dots, s_N^{*}\}$ encoded with the extended vocabulary, the model is trained to autoregressively predict each token by minimizing the negative log-likelihood:
\begin{equation}
  \mathcal{L}_{SFT}(\theta) = -\sum_{t=1}^{N} \log\, p_\theta(s_t^{*} \mid s_{<t}^{*},\, I_{tgt})
  \label{eq:sft}
\end{equation}
After this stage, the model can generate complete SVG code from an input image but remains unaware of the visual quality of its output, as no rendering feedback is provided during training.

\noindent\textbf{Stage 2: ReAct Cold-Start.}\quad
The Stage~1 model generates SVG from images but cannot interpret visual feedback or revise its own output. Because naturally occurring correction pairs are unavailable, we construct cold-start data through a rejection-sampling pipeline~\cite{yuan2023scaling}. The Stage~1 model is applied to the training images to produce draft SVGs, and each rendered prediction is scored against the target image via CLIP~\cite{DBLP:conf/icml/RadfordKHRGASAM21} similarity. A band-pass filter retains only drafts whose similarity falls within a moderate range, excluding near-perfect outputs that provide no learning signal and severely broken outputs whose correction would degenerate into full re-generation. An additional structural matching filter ensures that retained drafts share a compatible element topology with the ground-truth SVG, so that the supervision signal corresponds to local revision rather than wholesale rewriting. The ground-truth SVG of each selected sample then serves as the correction target. Filtering thresholds, selection statistics, and dataset size are reported in the supplementary material.

Each training instance is organized as a complete single-round ReAct episode:
\begin{equation}
  X = [I_{tgt},\; P_{init},\; S_{init},\; T_{call},\; I_{triplet},\; P_{react},\; S_{correct}^{*}]
  \label{eq:react_seq}
\end{equation}
where $P_{init}$ and $P_{react}$ denote the generation and correction prompts, $T_{call}$ is the tool-calling return from the rendering engine, $I_{triplet}$ is the visual feedback triplet defined in Section~\ref{sec:react}, and $S_{correct}^{*}$ is the ground-truth correction target. During training, the cross-entropy loss is computed on all model-generated segments, namely the initial draft $S_{init}$ and the corrected output $S_{correct}^{*}$, while all non-generated tokens including input images, text prompts, and tool-calling returns are masked. This joint supervision teaches the model to perform both open-loop generation and closed-loop correction within a unified training episode.

\noindent\textbf{Stage 3: Agentic Reinforcement Learning.}\quad
Supervised fine-tuning exposes the model only to curated correction examples, achieving surface-level behavioral alignment~\cite{DBLP:conf/nips/ZhouLX0SMMEYYZG23} without endowing the model with genuine correction capability that generalizes beyond the imitation distribution~\cite{gudibande2023false}. The model remains unable to explore diverse correction strategies or recover from novel generation errors. To close this gap, we turn to reinforcement learning from environmental feedback~\cite{DBLP:conf/nips/Ouyang0JAWMZASR22} and apply end-to-end optimization using Group Relative Policy Optimization (GRPO)~\cite{DBLP:journals/corr/abs-2402-03300}, a value-network-free alternative to PPO~\cite{schulman2017proximal} that estimates advantages from group-level reward statistics and has proven effective for scaling reasoning capabilities~\cite{guo2025deepseek}.

Given a target image and generation prompt as the initial context $q$, the policy $\pi_\theta$ autoregressively produces a group of $G$ complete agentic trajectories $\{o_i\}_{i=1}^{G}$. Each trajectory follows the ReAct episode structure defined in Eq.~\ref{eq:react_seq} and contains two model-generated segments---the initial SVG draft $S_{init}^{(i)}$ and the corrected output $S_{correct}^{(i)}$---interleaved with environment-provided content (text prompts, tool-calling returns, and the visual feedback triplet). Consistent with Stage~2, the policy gradient is computed exclusively over the model-generated token positions; all non-generated tokens are masked. The outcome reward $R_i$ (Section~\ref{sec:reward}) is determined solely by the final corrected output: $S_{correct}^{(i)}$ is extracted from each trajectory, rendered, and evaluated against the target image. The group-normalized advantage is $\hat{A}_i = (R_i - \mu_R) / \sigma_R$, and the training objective is:
\begin{equation}
  \mathcal{J}(\theta) = \mathbb{E}_{q,\, \{o_i\}} \!\left[ \frac{1}{G} \sum_{i=1}^{G} \min\!\left( \rho_i \hat{A}_i,\; \operatorname{clip}(\rho_i, 1{-}\epsilon, 1{+}\epsilon)\, \hat{A}_i \right) \right]
  \label{eq:grpo}
\end{equation}
where $\rho_i = \pi_\theta(o_i|q) / \pi_{\theta_{old}}(o_i|q)$ is the importance sampling ratio computed over the model-generated tokens in trajectory $o_i$. Following the analysis by Yu et al.~\cite{DBLP:journals/corr/abs-2503-14476} that KL regularization is unnecessary when the policy must diverge substantially from its initialization to acquire new capabilities, we remove the KL penalty to allow unconstrained exploration of correction strategies.

\subsection{Multi-dimensional Reward Design}
\label{sec:reward}

The outcome reward drives exploration in Stage~3. Each generated SVG $S_{pred}$ is rendered to produce a preview $I_{pred}$, and the reward is composed from three complementary components that jointly assess visual quality and code efficiency.

\noindent\textbf{Structural Fidelity Reward.}\quad
This component measures pixel-level similarity via mean squared error. An exponential decay with rate $\kappa$ maps the unbounded error to a reward in $[0, 1]$, yielding finer discrimination among near-correct outputs while saturating for large errors:
\begin{equation}
  R_{struct} = \exp\!\left(-\kappa \cdot \mathrm{MSE}(I_{pred},\, I_{tgt})\right)
  \label{eq:r_struct}
\end{equation}

\noindent\textbf{Semantic Perception Reward.}\quad
Pixel-level error is overly sensitive to minor spatial shifts that do not affect perceived quality. To capture high-level semantic alignment, we measure cosine similarity in the feature space of a pretrained DINOv2~\cite{DBLP:journals/tmlr/OquabDMVSKFHMEA24} encoder $f(\cdot)$ and apply a hard threshold $\tau$ that zeroes out severely misaligned outputs:
\begin{equation}
  R_{sem} = s \cdot \mathbb{I}[s \geq \tau],\quad s = \cos\bigl(f(I_{pred}),\, f(I_{tgt})\bigr)
  \label{eq:r_sem}
\end{equation}

\noindent\textbf{Code Efficiency Reward.}\quad
During RL exploration the model tends to stack redundant SVG elements that inflate code length without improving visual quality. Let $r = N_{pred} / N_{tgt}$ denote the ratio of the generated token length to the ground-truth length. We apply a cosine decay function:
\begin{equation}
  R_{eff} = \frac{1}{2}\!\left(1 + \cos\!\left(\pi \cdot \operatorname{clamp}\!\left(\frac{r - r_{tol}}{r_{max} - r_{tol}},\; 0,\; 1\right)\right)\right)
  \label{eq:r_eff}
\end{equation}
where $\operatorname{clamp}$ restricts the normalized ratio to $[0, 1]$, so that $R_{eff} = 1$ when $r \leq r_{tol}$ and $R_{eff} = 0$ when $r \geq r_{max}$, with smooth cosine decay in between.

\noindent\textbf{Reward Aggregation.}\quad
The three components are combined into a weighted visual reward $R_{vis} = w_1 R_{struct} + w_2 R_{sem} + w_3 R_{eff}$. To enforce adherence to the ReAct template and code renderability, we impose a hierarchical penalty conditioned on three failure indicators: core code absence ($\mathbb{I}_{miss}$), rendering crash ($\mathbb{I}_{render}$), and template format violation ($\mathbb{I}_{fmt}$):
\begin{equation}
  R_{total} =
  \begin{cases}
    \lambda_1 & \text{if } \mathbb{I}_{miss} = 1 \text{ or } \mathbb{I}_{render} = 1 \\[4pt]
    R_{vis} + \lambda_2 & \text{if } \mathbb{I}_{fmt} = 1 \\[4pt]
    R_{vis} & \text{otherwise}
  \end{cases}
  \label{eq:r_total}
\end{equation}
where $\lambda_1$ and $\lambda_2$ ($\lambda_1 \leq \lambda_2 < 0$) are fixed penalties. Fatal errors (missing code or rendering failure) override the visual reward entirely with the harshest penalty $\lambda_1$, while format violations reduce the reward by a constant offset $\lambda_2$, still allowing the model to receive partial credit for visual quality. 

\begin{table*}[!t]
  \centering
  \resizebox{0.985\textwidth}{!}{%
  \begin{tabular}{llccccccrr}
    \toprule
    \textbf{Method} & \textbf{Params} & \textbf{DINO}\,$\uparrow$ & \textbf{PSNR}\,$\uparrow$ & \textbf{CLIP-I2I}\,$\uparrow$ & \textbf{SSIM}\,$\uparrow$ & \textbf{LPIPS}\,$\downarrow$ & \textbf{MSE}\,$\downarrow$ & \textbf{\#Tok.} & \textbf{Time\,(s)} \\
    \midrule
    \rowcolor{groupgray}
    \multicolumn{10}{c}{\textbf{\textit{\small Optimization-based}}} \\
    DiffVG~\cite{li2020differentiable} & -- & 0.8419 & 32.48 & 0.9586 & 0.9539 & 0.1917 & 0.0008 & 268.43k & 109 \\
    LIVE~\cite{ma2022towards} & -- & 0.9053 & 33.93 & 0.9719 & 0.9804 & 0.0764 & 0.0006 & 72.59k & 402 \\
    \midrule
    \rowcolor{groupgray}
    \multicolumn{10}{c}{\textbf{\textit{\small General-purpose VLMs}}} \\
    Qwen3-VL-235B~\cite{Qwen3-VL} & 235B & 0.7690 & 6.58 & 0.8691 & 0.3279 & 0.4348 & 0.4463 & 8.65k & -- \\
    GPT-5.2~\cite{singh2025openai} & -- & \textbf{0.9247} & 11.92 & \textbf{0.9483} & 0.6114 & 0.2801 & 0.1228 & 849 & -- \\
    Gemini-3.1-Pro~\cite{google2026gemini} & -- & 0.9138 & 13.19 & \underline{0.9433} & 0.6564 & 0.2414 & 0.1205 & 1.14k & -- \\
    \midrule
    \rowcolor{groupgray}
    \multicolumn{10}{c}{\textbf{\textit{\small SVG-specialized models}}} \\
    StarVector~\cite{rodriguez2025starvector} & 8B & 0.7820 & 9.89 & 0.8830 & 0.4296 & 0.3329 & 0.4049 & 4.51k & 124 \\
    OmniSVG~\cite{yang2025omnisvg} & 8B & 0.8212 & 11.72 & 0.8838 & 0.6072 & 0.2784 & 0.2273 & 5.79k & 78 \\
    InternSVG~\cite{wang2025internsvg} & 8B & 0.8647 & 13.70 & 0.9098 & 0.6397 & \underline{0.2390} & 0.1980 & 8.34k & 34 \\
    \midrule
    \rowcolor{groupgray}
    \multicolumn{10}{c}{\textbf{\textit{\small RefineSVG (Ours)}}} \\
    Qwen2.5-VL-3B-Instruct & 3B & 0.6936 & 4.91 & 0.8365 & 0.3170 & 0.4998 & 0.4540 & 4.46k & 22 \\
    \quad +SVG-SFT & 3B & 0.7423 & 8.27 & 0.8648 & 0.3509 & 0.3911 & 0.4599 & 8.93k & 58 \\
    \quad +\textbf{RefineSVG} {\small(ours)} & 3B & 0.8872 & \underline{14.61} & 0.9171 & \underline{0.6638} & 0.2399 & \underline{0.0732} & 319 & 17 \\
    \rowcolor{deltbg}
    \quad\textbf{\textit{\small$\Delta$\,Gain}} & & \textcolor{delttext}{\textbf{\textit{\small$\uparrow$14.5}}} & \textcolor{delttext}{\textbf{\textit{\small$\uparrow$6.34}}} & \textcolor{delttext}{\textbf{\textit{\small$\uparrow$5.2}}} & \textcolor{delttext}{\textbf{\textit{\small$\uparrow$31.3}}} & \textcolor{delttext}{\textbf{\textit{\small$\downarrow$15.1}}} & \textcolor{delttext}{\textbf{\textit{\small$\downarrow$38.7}}} & \textcolor{delttext}{\textbf{\textit{\small$\downarrow$8.6k}}} & \\
    \midrule
    Qwen2.5-VL-7B-Instruct & 7B & 0.7315 & 6.64 & 0.8458 & 0.3862 & 0.4718 & 0.3477 & 3.97k & 34 \\
    \quad +SVG-SFT & 7B & 0.7725 & 10.14 & 0.8794 & 0.4097 & 0.3496 & 0.4121 & 8.91k & 100 \\
    \quad +\textbf{RefineSVG} {\small(ours)} & 7B & \underline{0.9207} & \textbf{15.86} & 0.9306 & \textbf{0.7114} & \textbf{0.1891} & \textbf{0.0603} & 634 & 27 \\
    \rowcolor{deltbg}
    \quad\textbf{\textit{\small$\Delta$\,Gain}} & & \textcolor{delttext}{\textbf{\textit{\small$\uparrow$14.8}}} & \textcolor{delttext}{\textbf{\textit{\small$\uparrow$5.72}}} & \textcolor{delttext}{\textbf{\textit{\small$\uparrow$5.1}}} & \textcolor{delttext}{\textbf{\textit{\small$\uparrow$30.2}}} & \textcolor{delttext}{\textbf{\textit{\small$\downarrow$16.1}}} & \textcolor{delttext}{\textbf{\textit{\small$\downarrow$35.2}}} & \textcolor{delttext}{\textbf{\textit{\small$\downarrow$8.3k}}} & \\
    \bottomrule
  \end{tabular}%
  }
  \caption{Quantitative comparison on SVG-Stack-1K. Among MLLM-based methods, the best results are in \textbf{bold} and second-best are \underline{underlined}. Optimization-based methods are listed for reference. \#Tok.\ denotes the average generated SVG token count. Time is measured in seconds per image on a single GPU. $\Delta$\,Gain reports the improvement of the full RefineSVG pipeline over the SVG-SFT baseline (Stage~1).}
  \label{tab:main}
\end{table*}

\section{Experiments}
\label{sec:experiments}

\subsection{Experimental Setup}

\noindent\textbf{Implementation Details.}\quad
We build RefineSVG on Qwen2.5-VL~\cite{Qwen2.5-VL} and report results for both the 3B and 7B variants, each extended with the SVG-oriented vocabulary described in Section~\ref{sec:tokenization}. All training is conducted on 32 NVIDIA H20 GPUs. The vision encoder and multimodal projector remain frozen throughout all stages; only the language model parameters are updated. In Stage~1, the model is trained for 3 epochs with a learning rate of $1 \times 10^{-5}$ and an effective batch size of 512. In Stage~2, we reduce the learning rate to $1 \times 10^{-6}$ and train for 2 epochs with a maximum context length of 24,576 tokens to accommodate the full ReAct episode. In Stage~3, we apply GRPO with $G = 8$ rollouts per prompt, an actor learning rate of $1 \times 10^{-6}$, a clipping ratio of $\epsilon = 0.2$, and a sampling temperature of 1.0. The reward weights are $w_1 = 0.3$, $w_2 = 0.5$, $w_3 = 0.2$, with $\kappa = 3.0$, $\tau = 0.3$, $r_{tol} = 1.5$, $r_{max} = 3.0$, $\lambda_1 = {-}0.5$, and $\lambda_2 = {-}0.3$. All stages use a cosine learning rate schedule.

\noindent\textbf{Datasets.}\quad
All training data originate from SVG-Stack~\cite{rodriguez2025starvector}. Stage~1 uses 1.52M quality-filtered image-SVG pairs rendered on a $256 \times 256$ canvas; Stage~2 uses 20K repair pairs obtained via rejection sampling (Section~\ref{sec:training}); Stage~3 uses a 35K candidate pool biased toward complex samples ($> 700$ tokens). For evaluation, we construct SVG-Stack-1K by stratified sampling from the SVG-Stack test split within the SVG-Bench~\cite{rodriguez2025starvector} suite, filtering out trivially simple instances to focus on samples with moderate to high structural complexity. All training data and evaluation benchmarks will be publicly released.

\noindent\textbf{Evaluation Metrics.}\quad
We use six complementary metrics. DINO and CLIP-I2I measure semantic similarity using DINOv2~\cite{DBLP:journals/tmlr/OquabDMVSKFHMEA24} and CLIP~\cite{DBLP:conf/icml/RadfordKHRGASAM21} image-level features, respectively. SSIM~\cite{DBLP:journals/tip/WangBSS04}, LPIPS~\cite{DBLP:conf/cvpr/ZhangIESW18}, PSNR, and MSE are standard image quality metrics. We additionally report the generated SVG token length (\#Tokens) to assess code efficiency. All our models are evaluated with a sampling temperature of 0.6; each result is averaged over three independent runs, with per-run variance reported separately.

\noindent\textbf{Baselines.}\quad
We compare against three categories of methods. (1)~Optimization-based: DiffVG~\cite{li2020differentiable} and LIVE~\cite{ma2022towards}. (2)~General-purpose VLMs: Qwen3-VL-235B~\cite{Qwen3-VL}, GPT-5.2~\cite{singh2025openai}, and Gemini-3.1-Pro~\cite{google2026gemini}, where the latter two are closed-source models evaluated via their official APIs. (3)~SVG-specialized models: StarVector-8B~\cite{rodriguez2025starvector}, OmniSVG-8B~\cite{yang2025omnisvg}, and InternSVG-8B~\cite{wang2025internsvg}. Complete training configurations, data construction pipelines, per-run variance, and inference prompts are detailed in the supplementary material.

\subsection{Main Results}
Table~\ref{tab:main} presents the quantitative comparison on SVG-Stack-1K. Among all MLLM-based methods, RefineSVG-7B achieves the best results on four of six quality metrics (PSNR, SSIM, LPIPS, MSE) and ranks second on DINO, establishing a new state of the art for open-source Image-to-SVG generation. Compared with the best competing SVG-specialized model InternSVG-8B, RefineSVG-7B improves DINO by +0.056, reduces LPIPS by 20.9\%, and lowers MSE by 69.5\%, demonstrating substantial gains in both semantic fidelity and pixel-level accuracy. Against the strongest closed-source model GPT-5.2, RefineSVG-7B trails only on two semantic similarity metrics, DINO ($-$0.004) and CLIP-I2I ($-$0.018), while surpassing it on all four remaining metrics by considerable margins, achieving +3.94 higher PSNR and 51\% lower MSE with only 7B parameters. Even the 3B variant already surpasses InternSVG-8B on five of six metrics despite having less than half the parameter count, confirming the effectiveness of the proposed training paradigm over pure scale.

Optimization-based methods (DiffVG, LIVE) attain the highest overall pixel-level fidelity through iterative gradient optimization, consistent with findings in concurrent work~\cite{wang2025internsvg}. However, this advantage comes at the cost of extreme SVG complexity and prohibitive latency: LIVE produces 72.59k tokens per image (over 114$\times$ that of RefineSVG-7B) and requires 402 seconds, yielding dense, non-editable path primitives that lack semantic structure.

A notable observation concerns the quality-efficiency tradeoff. Fewer tokens do not inherently imply better results: overly compact code may omit fine-grained details, while verbose code introduces redundant primitives without improving fidelity. Among MLLM-based methods, InternSVG uses 8.34k tokens and GPT-5.2 uses 849 tokens, yet both fall short of RefineSVG-7B on most quality metrics. The final corrected SVG produced by RefineSVG-7B averages only 634 tokens, a 14$\times$ reduction from its SFT counterpart (8.91k), while simultaneously improving all quality metrics. This indicates that the efficiency reward eliminates structural redundancy rather than blindly minimizing code length, yielding more information-dense SVG representations. Although the full ReAct episode generates both an initial draft and a corrected output, the combined token budget remains well below competing single-pass methods, translating directly into faster inference: RefineSVG-7B completes in 27 seconds, faster than all SVG-specialized baselines including InternSVG (34s), OmniSVG (78s), and StarVector (124s), and 3.7$\times$ faster than its single-pass SFT counterpart (100s). Both model scales exhibit consistent stage-wise improvements, confirming the scalability of the proposed framework.

\subsection{Ablation Studies}

To validate the contribution of each component, we conduct ablation experiments on both model scales (Table~\ref{tab:ablation}).

\begin{figure}[!htb]
  \centering
  \includegraphics[width=0.97\linewidth]{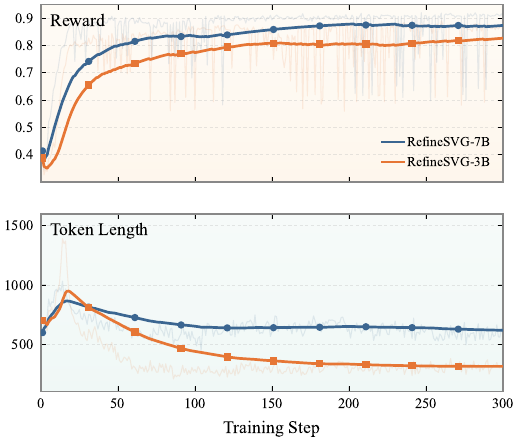}
  \caption{Stage~3 RL training dynamics. \textit{Top}: total reward steadily increases and converges for both model scales, with the 7B variant reaching a higher plateau. \textit{Bottom}: average generated token length drops rapidly from the SFT initialization and stabilizes, confirming that the efficiency reward effectively regularizes code redundancy without compromising generation quality.}
  \Description{Two training curves for the 3B and 7B models. The upper chart shows total reward increasing over training, and the lower chart shows generated token length decreasing and stabilizing.}
  \label{fig:training}
\end{figure}

Removing the cold-start stage (w/o Cold-Start) causes the largest degradation, with DINO dropping by 0.098 on the 3B model versus 0.032 on 7B. The disproportionate impact on the smaller model confirms that learning the ReAct correction paradigm from scratch via RL alone is prohibitively costly when model capacity is limited; cold-start supervision provides the behavioral scaffold that enables efficient RL convergence. Replacing the multi-dimensional Diff-Map with a plain rendered image (w/o Diff-Map) degrades quality on both scales, with the 7B model exhibiting a larger DINO decline ($-$0.019) than the 3B ($-$0.009). This scale-dependent pattern suggests that larger models are better positioned to exploit fine-grained spatial error signals, and removing this visual channel limits their correction precision.

\begin{table}[!hb]
  \centering
  \resizebox{\columnwidth}{!}{%
  \begin{tabular}{lccccc}
    \toprule
    \textbf{Variant} & \textbf{DINO}\,$\uparrow$ & \textbf{SSIM}\,$\uparrow$ & \textbf{LPIPS}\,$\downarrow$ & \textbf{MSE}\,$\downarrow$ & \textbf{\#Tok.} \\
    \midrule
    \rowcolor{groupgray}
    \multicolumn{6}{c}{\textbf{\textit{\small 3B}}} \\
    RefineSVG-3B (full) & \textbf{0.8872} & \textbf{0.6638} & \textbf{0.2399} & \textbf{0.0732} & 319 \\
    w/o Cold-Start & 0.7894 & 0.6254 & 0.2127 & 0.0957 & 806 \\
    w/o Diff-Map & 0.8785 & 0.6596 & 0.2420 & 0.0744 & 312 \\
    Direct RL & 0.8720 & 0.6618 & 0.2438 & 0.0895 & 347 \\
    \midrule
    \rowcolor{groupgray}
    \multicolumn{6}{c}{\textbf{\textit{\small 7B}}} \\
    RefineSVG-7B (full) & \textbf{0.9207} & \textbf{0.7114} & \textbf{0.1891} & \textbf{0.0603} & 634 \\
    w/o Cold-Start & 0.8886 & 0.7097 & 0.1888 & 0.0691 & 526 \\
    w/o Diff-Map & 0.9022 & 0.6921 & 0.2038 & 0.0662 & 464 \\
    Direct RL & 0.8798 & 0.7022 & 0.1929 & 0.0715 & 427 \\
    \bottomrule
  \end{tabular}%
  }
  \caption{Ablation study on SVG-Stack-1K. Each variant removes one component from the full RefineSVG pipeline.}
  \label{tab:ablation}
\end{table}

\noindent Applying RL directly after Stage~1 SFT without the agentic ReAct loop (Direct RL), an approach architecturally comparable to RLRF~\cite{rodriguez2025rendering}, consistently underperforms the full pipeline. The performance gap widens with model scale: DINO declines by 0.015 on 3B but 0.041 on 7B, indicating that the closed-loop agentic paradigm unlocks additional capacity that open-loop RL cannot exploit. This result validates that visual feedback at inference time, rather than reward-only feedback at training time, is essential for realizing the full potential of larger models. Notably, while some ablation variants produce fewer tokens (e.g., w/o Diff-Map yields 464 tokens on 7B vs.\ 634 for the full model), this comes at the cost of degraded fidelity, reinforcing that the efficiency reward serves as a regularizer against redundancy rather than a code-length minimizer.

Figure~\ref{fig:training} visualizes the Stage~3 training dynamics. The reward curves exhibit oscillatory but steadily ascending trajectories for both scales, with the 7B model converging to a higher plateau than 3B, reflecting a consistent scaling trend. The token length curves reveal an initial increase during early exploration, where the policy experiments with diverse correction strategies, followed by a gradual decline as the model discovers compact and efficient SVG representations. This convergence behavior is facilitated by the SVG-oriented vocabulary, which retains high-level semantic elements (e.g., \texttt{<rect>}, \texttt{<circle>}) rather than decomposing all geometry into low-level path commands, enabling the model to express complex structures with fewer tokens.

\begin{figure*}[!t]
  \centering
  \includegraphics[width=0.98\textwidth]{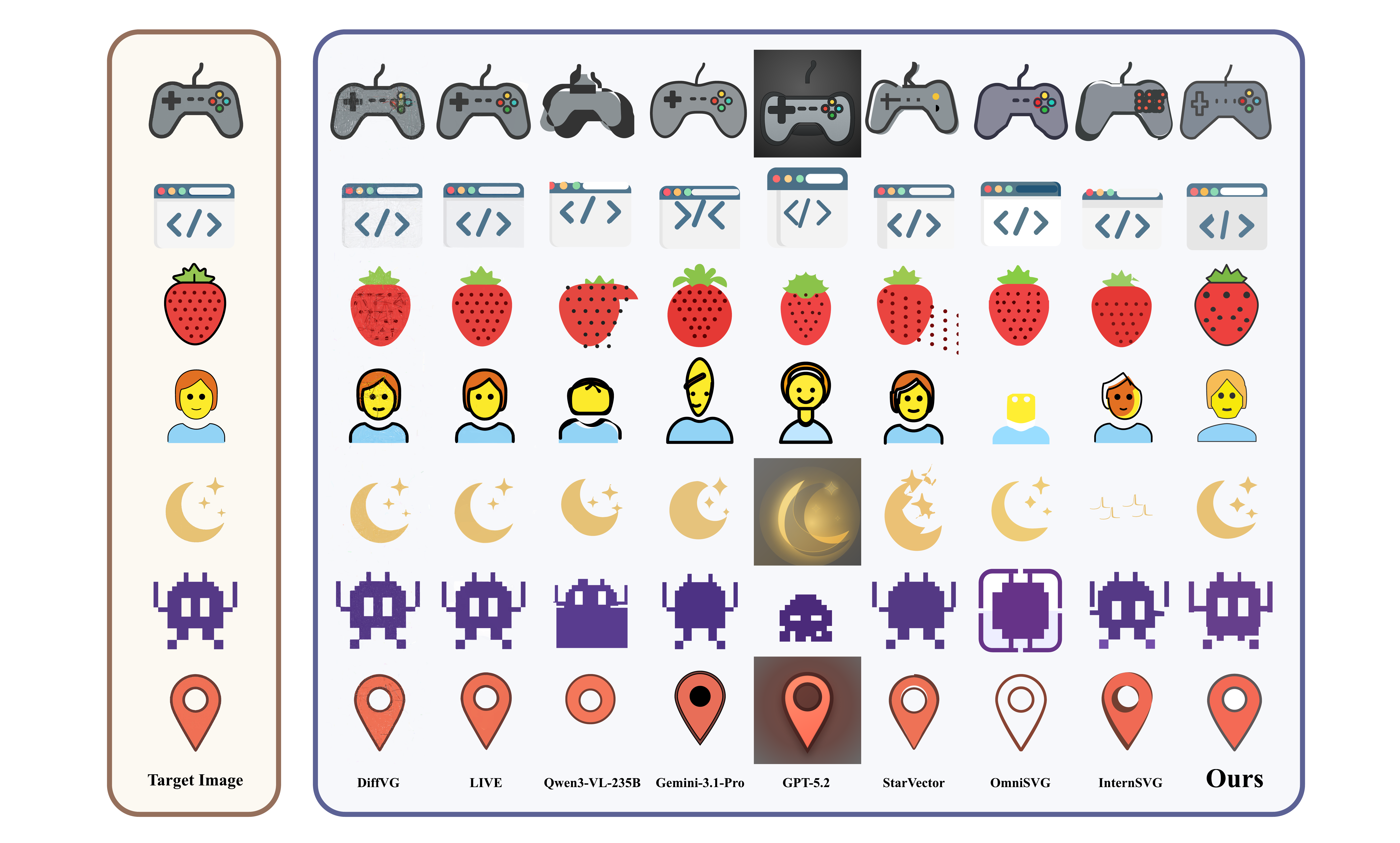}
  \caption{Qualitative comparison on out-of-distribution samples drawn from MMSVG-Illustration, MMSVGBench, and svg-emoji. RefineSVG-7B produces structurally coherent and visually faithful SVGs, preserving fine-grained details where competing methods exhibit geometric drift, color distortion, or structural omission.}
  \Description{A grid of out-of-distribution target images and SVG reconstructions from DiffVG, LIVE, three general-purpose vision-language models, three SVG-specialized models, and RefineSVG. RefineSVG outputs most closely preserve the target shapes, colors, and details.}
  \label{fig:qualitative}
\end{figure*}

\FloatBarrier
\subsection{Qualitative Results}

Figure~\ref{fig:qualitative} presents qualitative comparisons on out-of-distribution samples that are absent from the training data. Optimization-based methods (DiffVG, LIVE) reproduce pixel-level details but produce dense, non-semantic path primitives. General-purpose VLMs capture coarse layout but frequently hallucinate structural elements or distort fine geometry. Among SVG-specialized models, InternSVG achieves the closest results to ours but still exhibits noticeable color shifts and missing details in complex regions. RefineSVG-7B consistently generates clean, structurally coherent outputs with accurate color reproduction and complete element coverage, demonstrating strong generalization beyond the training distribution. Quantitatively, across SVG-Emoji, MMSVGBench, and MMSVG-Illustration, RefineSVG-7B outperforms all SVG-specialized baselines on the four reconstruction-oriented metrics (PSNR, SSIM, LPIPS, and MSE), while delivering competitive semantic similarity. Full DINO and CLIP-I2I results are provided in the supplementary material.

\FloatBarrier

\section{Conclusion}

We present RefineSVG, a single-step closed-loop visual feedback framework that elevates MLLMs from passive code generators to self-correcting visual agents for image-to-SVG generation. Through an SVG-oriented semantic vocabulary, a Diff-Map-guided ReAct mechanism, and a progressive training pipeline spanning SFT, rejection sampling for cold-start supervision, and agentic RL, RefineSVG achieves state-of-the-art results among MLLM-based methods on SVG-Stack-1K while generating substantially shorter and more efficient SVG code than competing approaches. The current framework performs a single correction round per image. Extending it to multi-round refinement remains an open challenge because constructing reliable cold-start supervision for multi-turn correction is considerably more difficult. Developing adaptive stopping criteria and curricula for deciding when further refinement is beneficial, as well as scaling to higher-resolution canvases, are promising directions. This paradigm may also generalize to multimodal structured code generation beyond SVG.

\begin{acks}
This work was supported in part by the National Natural Science Foundation of China (Grant 62572327) and the Guangdong Basic and Applied Basic Research Foundation (Grant 2025A1515010260). Author contributions: Shaobo Liu conceived the core method, designed and implemented the full training pipeline, and wrote the manuscript. Shuaishuai Zhou conducted the baseline experiments and evaluation across multiple models and contributed to technical discussions. Yan Zhan designed the main method figure and data-processing visualizations and reviewed the final manuscript.
\end{acks}

\balance
\bibliographystyle{ACM-Reference-Format}
\bibliography{references}

\clearpage
\nobalance
\includepdf[pages=-,pagecommand={\thispagestyle{empty}},fitpaper=true]{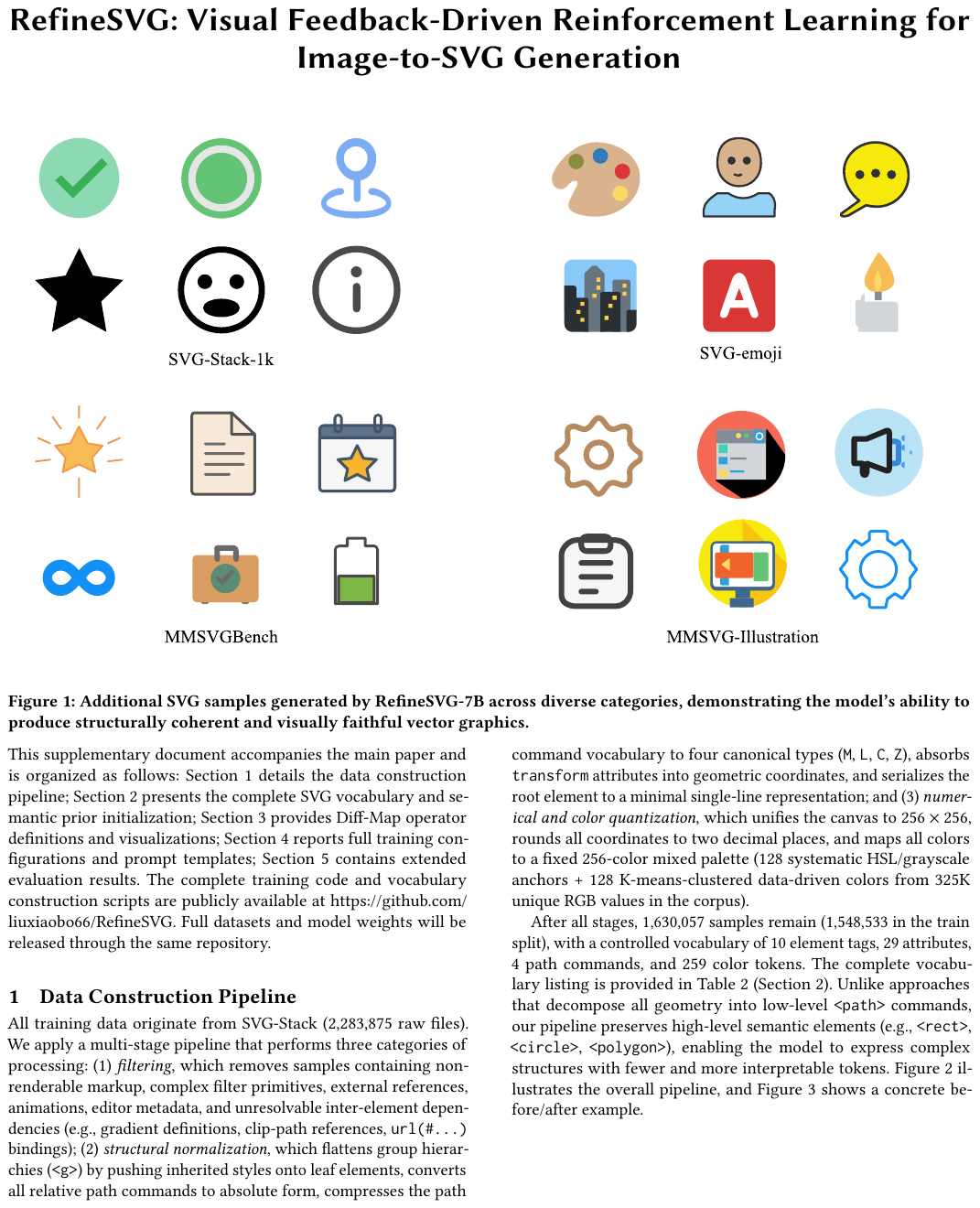}

\end{document}